\documentclass[runningheads]{llncs}

\usepackage[T1]{fontenc}

\usepackage{graphicx}
\graphicspath{{annexes/}}

\usepackage{tabularray}
\UseTblrLibrary{booktabs}

\usepackage{multirow}
\usepackage{bm}
\usepackage{array}
\usepackage{amsmath}

\usepackage{lmodern}

\usepackage{orcidlink}

\usepackage{tikz}
\usetikzlibrary{arrows, external, positioning}

\usepackage{subcaption}

\usepackage{hyperref}
\usepackage{color}

\begin{document}


\title{Benefits of Low-Cost Bio-Inspiration in the Age of Overparametrization}
\titlerunning{Benefits of Low-Cost Bio-Inspiration}

\author{Kevin Godin-Dubois\inst{1}\orcidlink{0009-0002-6033-3555} \and
Anil Yaman\inst{1}\orcidlink{0000-0003-1379-3778} \and
Anna V. Kononova\inst{2}\orcidlink{0000-0002-4138-7024}}

\authorrunning{K. Godin-Dubois et al.}

\institute{
 Vrije Universiteit Amsterdam, Netherlands \\ \email{k.j.m.godin-dubois@vu.nl} \and
 LIACS, Leiden Universiteit, Netherlands
}
\maketitle              
\begin{abstract}
While Central Pattern Generators (CPGs) and Multi-Layer Perceptrons (MLP) are widely used paradigms in robot control, few systematic studies have been performed on the relative merits of large parameter spaces in highly constrained settings.
As opposed to traditional Machine Learning contexts, our input and output spaces are small and performance is bounded thus having more parameters may actively hinder the learning process instead of empowering it.
To empirically measure this, we submit a given robot morphology, with limited proprioceptive capabilities, to controller optimisation under two bio-inspired paradigms (CPGs and MLPs) with evolutionary- and reinforcement- trainer protocols.
By varying parameter spaces across multiple reward functions, we demonstrate that shallow MLPs and densely connected CPGs result in better performance when compared to deeper MLPs or Actor-Critic architectures.
To account for the relationship between said performance and the number of parameters, we introduce a Parameter Impact metric which showcases diminishing returns for MLPs but not for CPGs.
Taken together these results demonstrate, on a fixed quadrupedal morphology, the benefits of integrating prior bias when considering locomotion tasks with simple hinge actuators.

\end{abstract}

\section{Introduction}

In Evolutionary Robotics (ER), morphologies and controllers are jointly optimised in a mutually dependent fashion; thus, every change in the former has to be accommodated by the latter \cite{Cheney2016}.
Unlike quadrupedal locomotion on high-end robotics platforms \cite{Owaki2017,Tan2018,Hwangbo2019,liu_reinforcement_2023,bellegarda_cpg-rl_2022,Raffin2024,bellegarda_visual_2024}, where the morphology is fixed to allow real-world deployment on existing machines, researchers have fewer assumptions to start from, including the lack of well defined limbs and optimal gaits.
Furthermore, deploying such evolved body plans in the lab typically requires using low-cost modular components \cite{CussatBlanc2014,Jelisavcic2017,Angus2023} with limited proprioceptive capabilities.
While practitioners have had success with many controller types including Central Pattern Generators (CPGs) \cite{Lan2021b,VanDiggelen2021} or Multi-Layer Perceptrons (MLPs) \cite{Diggelen2024,luo_comparison_2023}, there are no clear guidelines on how to design a controller architecture for such robots.
To fill this knowledge gap, this work\footnote{Sources: \url{https://github.com/kgd-al/apets-ariel/releases/tag/CPG_v_MLP}} explores the impact of algorithmic and architectural choices on the performance of evolved gaits in a directed locomotion task.

The robot used in this study is a specific instance of modular robots \cite{Eiben2015a} composed of a central control module, 8 articulations and 8 structural blocks.
Each articulation possesses a single degree of freedom and provides a read out of their current position in one-dimensional local coordinates.
The controller thus works with an 8-dimensional input and output space, the latter being on par with existing benchmarks, while the former is much smaller.
As a comparison, a similar morphology widely used as a Reinforcement Learning (RL) benchmark \cite{Schulman2018} has the same number of actuators but relies on 105 observations, which may lead to a wider reality gap \cite{Tan2018}.
Thus, in such a constrained space, the typical insights from Machine Learning (ML) that overparametrization leads to better results \cite{Neyshabur2018,Jiang2019,Gastpar2023} may not hold.

To thoroughly investigate this, we consider two training algorithms, namely the Covariance Matrix Adaptation Evolutionary Strategy (CMA-ES) and Proximal Policy Optimisation (PPO), and, through systemic comparisons, contribute the following:
\begin{itemize}
 \item CPGs and MLPs are capable of similar levels of performance but the former is more parameter-efficient than the latter;
 \item CMA-ES performs overall better than PPO when considering efficiency;
 \item Both CPGs and MLPs are sensitive to reward functions with an impact on both performance and efficiency.
\end{itemize}

\section{Literature}

Designing controllers for autonomous robots has been extensively studied in numerous fields including, but far from limited to, Evolutionary Computation~\cite{back2023evolutionary} and Reinforcement Learning.
In this work, we focus on these two specific research fields as both have been successfully used in an Evolutionary Robotics (ER) context \cite{Eiben2015a}.
Indeed, simultaneously optimising both a robot's morphology and controller leads to many, currently unsolved, problems \cite{Cheney2016} stemming from potential mismatches and short-term performance loss.
However, in ER, CPGs and MLPs stand out as the two major design principles for generic controllers with many instances of the former \cite{Lan2021b,VanDiggelen2021,Jelisavcic2019,luo_comparison_2023,Luo2022} and latter \cite{Diggelen2024,luo_comparison_2023} being used both with and without morphological coevolution.

On the one hand, CPGs have been shown to be very efficient at producing biologically plausible motion patterns for e.g. snakes \cite{liu_reinforcement_2023}, quadrupeds \cite{bellegarda_cpg-rl_2022,Watanabe2025,bellegarda_visual_2024}, hexapods \cite{wang_central_2017} or swimmers \cite{Ijspeert1999,ijspeert_central_2008}.
On the other hand, MLPs have been widely used for all sorts of morphologies, in both EC \cite{GodinDubois2023,GodinDubois2024b,Baldominos2020,Stanley2019} and RL \cite{Tomilin2022,GodinDubois2025d,tsounis_deepgait_2019,OpenAI2019} contexts.
Nonetheless, while both methodologies seem similarly competent at providing viable solutions to specific locomotion optimisation tasks, they have seldom been compared in such a restricted setting.
While one such work has been done in \cite{luo_comparison_2023}, the neural architectures used many inputs and only considered a single configuration for CPGs networks.
Furthermore, reproducibility is hampered by the reliance on custom implementation, whereas this work solely uses off-the-shelf libraries with minimal changes.
Another comparison has been undertaken in \cite{VanDiggelen2021}, which focuses only on CPGs with a single architecture and uses two gradient-free methods.

While CPGs have usually relied on specific architectural designs, MLPs have been used with a much wider range of configurations in both manual and automated manners \cite{Zoph2017a,Brock2017,Liu2018,White2023,Hegde2024}.
Current trends in Reinforcement Learning point towards using more and more parameters thanks to the observed benefits of power laws \cite{Zhang2017a}, however this approach is far from frugal and may not even be necessary \cite{Wong2026}.
Interestingly, the wider Machine Learning community has recently re-invented the concept of structural bias~\cite{Kononova2015} drawing attention to the fact that parameters count is only important alongside a given optimiser~\cite{Pascanu2025,Mohan2025}.
Overparametrization, i.e. having more parameters than the number of training points \cite{Neyshabur2018}, while performing well in many ML tasks, is poorly understood on a theoretical level \cite{Jiang2019,Gastpar2023}.
In continuous control, especially for locomotion, simpler solutions have also shown robust results \cite{Rajeswaran2017,Tan2018,Hegde2023}.
We contribute to this growing body of knowledge on more frugal approaches by providing a systematic exploration of the impact of parameter space on both performance and efficiency.

\begin{figure}[t]
 \centering
 \includegraphics[width=.25\textwidth]{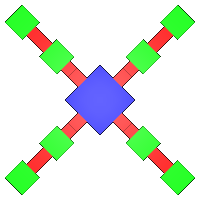}
 \caption{Robot ``spider'' morphology with 8 actuated hinges (red), 8 structural bricks (green) and a central controller (blue)}
 \label{fig:spider_body}
\end{figure}

\section{Methods}

To study the relative benefits of specific neural architectures, we leverage the ARIEL (formerly known as Revolve \cite{Stuurman2024zd}) platform, which has been extensively used with both CPGs \cite{Lan2021b,VanDiggelen2021,Jelisavcic2019,luo_comparison_2023,Luo2022} and MLPs \cite{Diggelen2024,luo_comparison_2023}.
In this framework, robots are composed of rigid morphologies, with either passive (core, brick) or active (hinges) components.
The core corresponds to the real-world central processing unit containing a Raspberry Pi hardware for generic control, as well as batteries, while bricks only serve as structural components.
Hinges are based on DSS-M15 actuators augmented with a positional output.
This library relies on MuJoCo \cite{Todorov2012} to handle the physical interactions in simulation to reach a high degree of robustness and reproducibility.

In this work, we focus on a traditional, so-called ``spider'', morphology illustrated in \autoref{fig:spider_body}.
This robot has 8 active hinges, both horizontal (``hip'' joints) and vertical (``knee'' joints), giving it a broad range of motion approaching traditional quadrupedal locomotion (where hip joints would have 2 degrees of freedom).
It follows that the observation and action spaces have dimension 8 with values clipped in $[-1,1]$.

\begin{figure}[t]
 \centering
 \begin{subfigure}{.3\textwidth}
  \centering
  \begin{tikzpicture}[>=stealth']
   \node (yi) [circle, draw] {$y_i$};
   \node (xi) [right=of yi, circle, draw] {$x_i$};
   \draw [->] (yi) to [bend left] node [pos=.5, anchor=south] {\small$w_i$} (xi);
   \draw [->] (xi) to [bend left] node [pos=.5, anchor=south] {\small$-w_i$} (yi) ;

   \node (yj) [below=of yi, circle, draw] {$y_j$};
   \node (xj) [right=of yj, circle, draw] {$x_j$};
   \draw [->] (yj) to [bend left] node [pos=.5, anchor=north] {\small$w_j$} (xj);
   \draw [->] (xj) to [bend left] node [pos=.5, anchor=north] {\small$-w_j$} (yj) ;
   
   \draw [->] (xi) to [bend left] node [pos=.5, inner sep=.5pt, fill=white] {\small$w_{ij}$} (xj) ;
   \draw [->] (xj) to [bend left] node [pos=.5, inner sep=.5pt, fill=white] {\small$-w_{ij}$} (xi) ;
  \end{tikzpicture}
  \caption{Paired CPGs}
  \label{fig:cpg:pair}
 \end{subfigure}
 \begin{subfigure}{.3\textwidth}
  \centering
  \tikzsetnextfilename{cpg-architectures}
  \begin{tikzpicture}[scale=.9]
   \tikzset{
    0/.style={solid, red},
    2/.style={red!66!black, dash pattern=on 2pt off 2pt},
    4/.style={red!33!black, dash pattern=on 2pt off 2pt on \pgflinewidth off 2pt},
    6/.style={black, dash pattern=on 2pt off 2pt on \pgflinewidth off 2pt on \pgflinewidth off 2pt},
   }
  
   \foreach \a in {45, 135, 225, 315} {
    \foreach \l in {1, 2} {
     \node (\a_\l) at (\a:\l) [circle, draw, 0] {};
    }
   }

   \foreach \l/\r in {
    45_1/45_2, 135_1/135_2, 225_1/225_2, 315_1/315_2,
    45_1/135_1, 135_1/225_1, 225_1/315_1, 315_1/45_1%
   } { \draw [2] (\l) to (\r); }

   \foreach \l/\r in {
    45_2/135_1, 45_2/315_1, 135_2/45_1, 135_2/225_1,
    225_2/135_1, 225_2/315_1, 315_2/225_1, 315_2/45_1,
    45_1/225_1, 135_1/315_1%
   } { \draw [4] (\l) to (\r); }

   \foreach \l/\r in {
    45_2/135_2, 135_2/225_2, 225_2/315_2, 315_2/45_2%
   } { \draw [6] (\l) to (\r); }

   \draw [6] (45_2) to [out=187.5, in=82.5, looseness=1.75] (225_2);
   \draw [6] (135_2) to [out=277.5, in=172.5, looseness=1.75] (315_2);

   \node at (0, 2.5) {Neighbourhood};
   \foreach \n [count=\i] in {0,2,4,6} {
    \draw [\n] (-3+\i, 2) -- ++(.5, 0) node [pos=1, anchor=west, black] {\n};
   }
  \end{tikzpicture}
  \caption{CPG Network}
  \label{fig:cpg:network}
 \end{subfigure}
 \begin{subfigure}{.3\textwidth}
  \begin{align*}
   \dot y &= -w_i x_i \\
   \dot x &= w_i y_i + \sum\limits_{j \in \mathcal{N}_i} w_{ij} x_j,
  \end{align*}
  \caption{Equations system}
  \label{fig:cpg:equations}
 \end{subfigure} 
 \caption{Neighbourhood configurations for a CPG network.}
\end{figure}
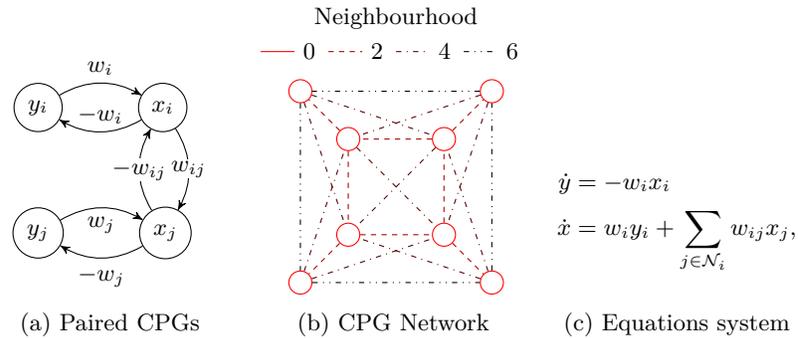

\subsection{CPG controller}

We rely on Central Pattern Generators, as implemented in similar studies on the legacy version of the platform \cite{Lan2021b,VanDiggelen2021,Jelisavcic2019,luo_comparison_2023,Luo2022}.
These take the form of coupled neurons $x$ and $y$ with recurrent connections as illustrated in \autoref{fig:cpg:pair}.
Traditional practices use the same weight for both connections between $x$ and $y$ but set one as excitatory and the other as inhibitory.
This results in periodical and bounded dynamics, ideally suited for rhythmic patterns such as locomotion.
However, as these neurons take no external inputs, the functions thereby produced would have very limited complexity.
To alleviate this, CPGs are usually connected together to form a network where the external node ($x$) also receives values from neighbouring CPGs.
Here, the weights are also symmetrically assigned with the connection weight $w_{ji}$ being equal to $-w_{ij}$.
The final form of the equation system for one neuron is given by \autoref{fig:cpg:equations}, where $\mathcal{N}_i$ is the neighbourhood of CPG $i$.
Usually, two neurons are said to be neighbours if their distance in the morphological tree is less than or equal to two \cite{Luo2022}.
However, in this work, we consider neighbourhood distance as a hyperparameter of the CPG network and vary it across its range, as illustrated on \autoref{fig:cpg:network}.

The simplest network, thereafter designed $c_0$, has no neighbourhood and thus only uses the first terms in \autoref{fig:cpg:equations}.
This, in turn, implies that the optimisation process will have only 8 parameters to work with, reducing the search space but not necessarily the quality of the solutions.
The remaining sub-architectures ($c_2$, $c_4$, $c_6$) have respective neighbourhood ranges of 2, 4 and 6.
As only the distance between hinges is considered, $c_0$ and $c_1$ are functionally equivalent as no hinges are directly connected to one another in the spider morphology.
Thus, the largest parameter space is 36 for the fully connected $c_6$, using all feedback pathways depicted on \autoref{fig:cpg:network}.

\begin{figure}[t]
 \tikzset{
  IO/.style={solid, red, circle, draw, inner sep=1pt},
  layer/.style={draw, minimum width=.5cm, minimum height=1cm},
  1/.style={layer, dash pattern=on 2pt off 2pt},
  2/.style={layer, dash pattern=on 2pt off 2pt on \pgflinewidth off 2pt},
  weight/.style={thin, black!25!white},
 }
 \def\angles{140, 160, ..., 220}
 
 \begin{subfigure}{.32\textwidth}
  \centering
  \begin{tikzpicture}[>=stealth']
   \foreach \i in {0, ..., 7} {
    \node (I\i) at (0, .1*\i) [IO] {};
    \node (O\i) at (1.5, .1*\i) [IO] {};
   }
   \draw [<->] ([xshift=-.5em]I0.west) -- ([xshift=-.5em]I7.west) node [pos=.5, anchor=east] {8};
   \draw [<->] ([xshift=+.5em]O0.east) -- ([xshift=+.5em]O7.east) node [pos=.5, anchor=west] {8};
   \foreach \i in {0, ..., 7} {
    \foreach \o in {0, ..., 7} {
     \draw [weight] (I\i) -- (O\o);
    }
   }
  \end{tikzpicture}
  \caption{Perceptron $m^0_0$}
  \label{fig:ann:mlp0}
 \end{subfigure}
 \begin{subfigure}{.32\textwidth}
  \centering
  \begin{tikzpicture}[>=stealth']
   \foreach \i in {0, ..., 7} {
     \node (I\i) at (0, .1*\i) [IO] {};
     \node (O\i) at (2.5, .1*\i) [IO] {};
    }
    \node [1] (1) at (1.25, .35) {};
    \draw [<->] (1.north) -- (1.south) node [pos=.5, fill=white] {w};
    \foreach \i in {0, ..., 7} {
     \foreach \a in \angles {
      \draw [weight] (1.\a) -- (I\i);
      \pgfmathsetmacro{\b}{180-\a}
      \draw [weight] (1.\b) -- (O\i);
     }
    }
  \end{tikzpicture}
  \caption{Single hidden layer $m^1_w$}
  \label{fig:ann:mlp1}
 \end{subfigure}
 \begin{subfigure}{.32\textwidth}
  \centering
  \begin{tikzpicture}[>=stealth']
    \foreach \i in {0, ..., 7} {
     \node (I\i) at (0, .1*\i) [IO] {};
     \node (O\i) at (3, .1*\i) [IO] {};
    }
    \node [1] (1) at (1, .35) {};
    \draw [<->] (1.north) -- (1.south) node [pos=.5, fill=white] {w};
    \node [2] (2) at (2, .35) {};
    \draw [<->] (2.north) -- (2.south) node [pos=.5, fill=white] {w};
    \foreach \i in {0, ..., 7} {
     \foreach \a in \angles {
      \draw [weight] (1.\a) -- (I\i);
      \pgfmathsetmacro{\b}{180-\a}
      \draw [weight] (2.\b) -- (O\i);
     }
    }
    \foreach \a in \angles {
     \foreach \b in \angles {
      \pgfmathsetmacro{\c}{180-\b}
      \draw [weight] (1.\c) -- (2.\a);
     }
    }
  \end{tikzpicture}
  \caption{2 hidden layers $m^2_w$}
  \label{fig:ann:mlp2}
 \end{subfigure}
 \caption{ANN architectures with $w$ denoting the width of hidden layers.}
 \label{fig:mlp}
\end{figure}
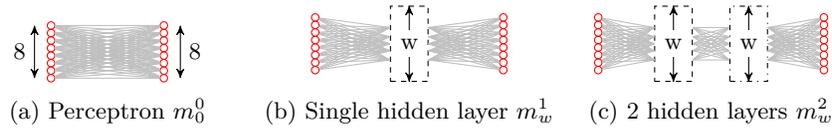

\subsection{ANN controller}

\begin{table}[b]
 \caption{Parameter spaces for CPGs with neighbourhood range $\mathcal{N}$ and for MLPs with width $w$ and depth $d$. Note that PPO has more parameters because of its actor and critic components.}
 \label{tab:parameters}
 \begin{subtable}{.29\textwidth}
  \caption{CPG}
  \label{tab:parameters:cpg}
  \begin{tblr}{lcc}
  \toprule
   & $\mathcal{N}$ & Parameters \\
  \midrule
   $c_0$ & 0 & 8 \\
   $c_2$ & 2 & 18 \\
   $c_4$ & 4 & 30 \\
   $c_6$ & 6 & 36 \\
  \bottomrule
  \end{tblr}
 \end{subtable}
 \begin{subtable}{.69\textwidth}
  \caption{MLP}
  \label{tab:parameters:mlp}
  \begin{tblr}{lcccc}
  \toprule
   \SetCell[r=2]{l} Name &
     \SetCell[r=2]{c} $d$ &
       \SetCell[r=2]{c} $w$ &
        \SetCell[c=2]{c} Parameters \\
  \cmidrule[lr]{4-6}
    & & & CMA-ES & PPO \\
  \midrule
   $m^0_0$ & 0 & N/A & 72 & 89 \\
   $m^1_w$ & 1 & $n$ & $17w+8$ & $27w+17$ \\
   $m^2_w$ & 2 & $n$ & $(18+w)w+8$ & $(29+2w)w+17$ \\
  \bottomrule
  \end{tblr}
 \end{subtable}
\end{table}

In a similar fashion to the CPG architecture, we rely on out-of-the-box implementations for the Artificial Neural Network controllers.
As will be detailed later on, we use Stable Baselines3 \cite{stable-baselines3} for our RL training, which also provides a default network architecture.
In its baseline configuration, ANNs used in conjunction with PPO \cite{Schulman2017} have two layers of 64 neurons with hyperbolic tangent activation function except for the output layer.
There, the network outputs are given by a Gaussian distribution, during training, which simplifies to the identity function during deterministic evaluation.

While, compared to CPGs, the hyperparameter space here is much richer, we focus our investigation on the direct impact of the optimised parameters.
Only the depth ($d$) and width ($w$) of networks are varied with $d \in \{0,1,2\}$ and $w \in \{1, 2, 4, 8, 16, 32, 64, 128\}$.
Furthermore, we denote a Multi-Layer Perceptron with a given depth and width as $m^d_w$.
Note that this labelling is expedient but somewhat imprecise, when referring to the sub-architecture $m^0_0$ as these are functionally perceptrons, as can be seen in \autoref{fig:ann:mlp0}.

For $d \in \{1, 2\}$, the ANN can accurately be called an MLP with number of parameters summarised in \autoref{tab:parameters}.
As will be discussed in \autoref{sec:model:trainers}, the major difference between our optimisation algorithms is the use of an actor-critic policy with Reinforcement Learning which exponentially increases the parameter space's size.

\subsection{Fitness/Reward functions}

In addition to varying hyperparameters of both CPGs and MLPs, we also look at the potential impact of the training regime, that is the function defining performance.
In this work, we use three methods that favour different gaits and learning trajectories, as discussed below.

\subsubsection*{Speed}
\def\rewS{$R_s$}
The most straightforward reward function is the speed along the x-axis.
Given a robot at x positions $x(t)$ and $x(t+1)$ at times $t$ and $t+1$, respectively, we define its reward \rewS{} as:
\begin{equation}
 R_s(t) = x(t+1) - x(t).
\end{equation}

The corresponding fitness function, for use with evolutionary algorithms, is obtained by summing over all time steps.
Thus, for a simulation of T seconds, with sampling rate of 20 Hz $R_s = \sum_{0 \leq t < T} R_s(t)$.
The same methodology applies to the subsequent reward/fitness function pairs and will thus not be repeated.

\subsubsection*{Gymnasium}
\def\rewG{$R_g$}
is a collection of environments actively used for benchmarking Reinforcement Learning algorithms \cite{Towers2024}.
Through independent convergence of designs, the ant\footnote{\url{https://gymnasium.farama.org/environments/mujoco/ant/}} morphology of this benchmark collection happens to be very similar to ARIEL's spider \cite{Lan2021b}.
Thus, with minimal tweaking, we ported the reward function used in this environment to train controllers that, theoretically, use hinges more frugally.
Given, at time $t$, a robot with forward velocity $V_x(t)$, applying a control signal $\vec S(t)$ to its hinges and receiving contact forces $\vec F(t)$, its so-called gymnasium reward \rewG$(t)$ is defined as:

\begin{equation}
 \label{eq:reward:gymnasium}
 R_g(t) = V_x(t) - .5 ||S(t)||_2 - .0005 ||F(t)||_2, 
\end{equation}

where $||.||_2$ represents the Euclidean distance (L2 norm).
Succinctly, this function promotes forward locomotion, as does \rewS{}, but also penalises large control forces, thereby favouring frugal locomotion, as well as contact forces, i.e. not hitting the ground hard.

\subsubsection*{Kernels-based}
\def\rewK{$R_k$}
The third and final reward function is based on Radial Basis Function (RBF) kernels as they provide a very distinct reward landscape.
Given that the robot, at time $t$, has velocity $V(t) = (V_x(t), V_y(t), V_z(t))$ and elevation $z(t)$, we define its kernel-based reward \rewK as:
\begin{equation}%
 \label{eq:reward:kernels}%
 \begin{aligned}%
  K(v, \hat v, c) &= e^{-c(v-\hat v)^2}\\ 
  R_k(t) &= \frac{5}{8} K(V_x(t), 0.5, 25) + \frac{1}{8} K(V_y(t), 0, 5) \\
         &+ \frac{1}{8} K(V_z(t), 0, 5) + \frac{1}{8} K(z(t), 0.2, .002), 
 \end{aligned}
\end{equation}
where $K(v, \hat v, c)$ defines a radial basis function with target value $\hat v$ and slope $c$.
With this function, the robot is thus encouraged to reach the target velocity (50 cm/s) with no lateral or vertical speed.
Furthermore, the body is expected to stay at an elevation of 20 cm, corresponding to 10 cm of distance between the bottom of the core and the ground which penalises crawling gaits.

\subsection{Optimisation setup}
\label{sec:model:trainers}

The final variable of adjustment investigated in this work is the optimisation algorithms.
On the one hand, we use the Covariance Matrix Adaptation Evolutionary Strategy \cite{Hansen1996,hansen2019pycma}, a powerful gradient-free optimisation method successfully used in conjunction with this morphological space \cite{Diggelen2024}.
Following usual practice with the former version of the framework, we keep all hyperparameters to their default configuration as detailed in the supplementary material \cite{kevin_2026_19633625}.
As the method is gradient-free, we used it for both CPGs and MLPs optimisation forming two broad groups which we will reference as \textbf{CMA/CPG} and \textbf{CMA/MLP} from now on.

Individuals were evaluated for 10 seconds each, with a control frequency of 20~Hz resulting in 200 time steps per simulation.
Every run consisted of 10'000 such evaluations with 10 replicates for every condition.
However, as we relied on an off-the-shelf naive implementation, the algorithm was unable to successfully explore large parameter spaces leading to prohibitive matrix size for the largest MLP architectures (more than 5Gb for $m^2_{128}$) and premature convergence.
This resulted in the exclusion of $m^1_{128}$, $m^2_{64}$, $m^2_{128}$ to limit resources wastage although some of the included runs still exhibited this form of premature convergence.

On the other hand, we also used Proximal Policy Optimisation \cite{Schulman2017} to train MLPs via a form of gradient descent as also used within this framework \cite{luo_comparison_2023}.
As this method is inefficient with recurrent networks, no off-the-shelf solutions existed for CPGs leading to only one group for this trainer: \textbf{PPO/MLP}.
The algorithm does not rely on episodic learning but, instead, works at the time step level.
Thus to ensure fair comparison with the CMA group we let PPO sample for 2'000'000 time steps, as is done in similar settings \cite{Milano2022,Wong2026}.
Environments are reset every 10 seconds (200 timesteps) leading to comparable episodes.
With respect to the hyperparameters, we use those recommended for gymnasium ant\footnote{\url{https://github.com/araffin/rl-baselines-zoo/blob/master/hyperparams/ppo2.yml}} which, as already stated, has similar morphological features.

\section{Results}

\subsection{Relative performance}

\begin{figure}
 \foreach \i/\c in {1, 3/\rewS, 4/\rewG, 2/\rewK} {
  \begin{subfigure}{.99\textwidth}%
   \centering%
   \includegraphics[width=\textwidth, page=\i]{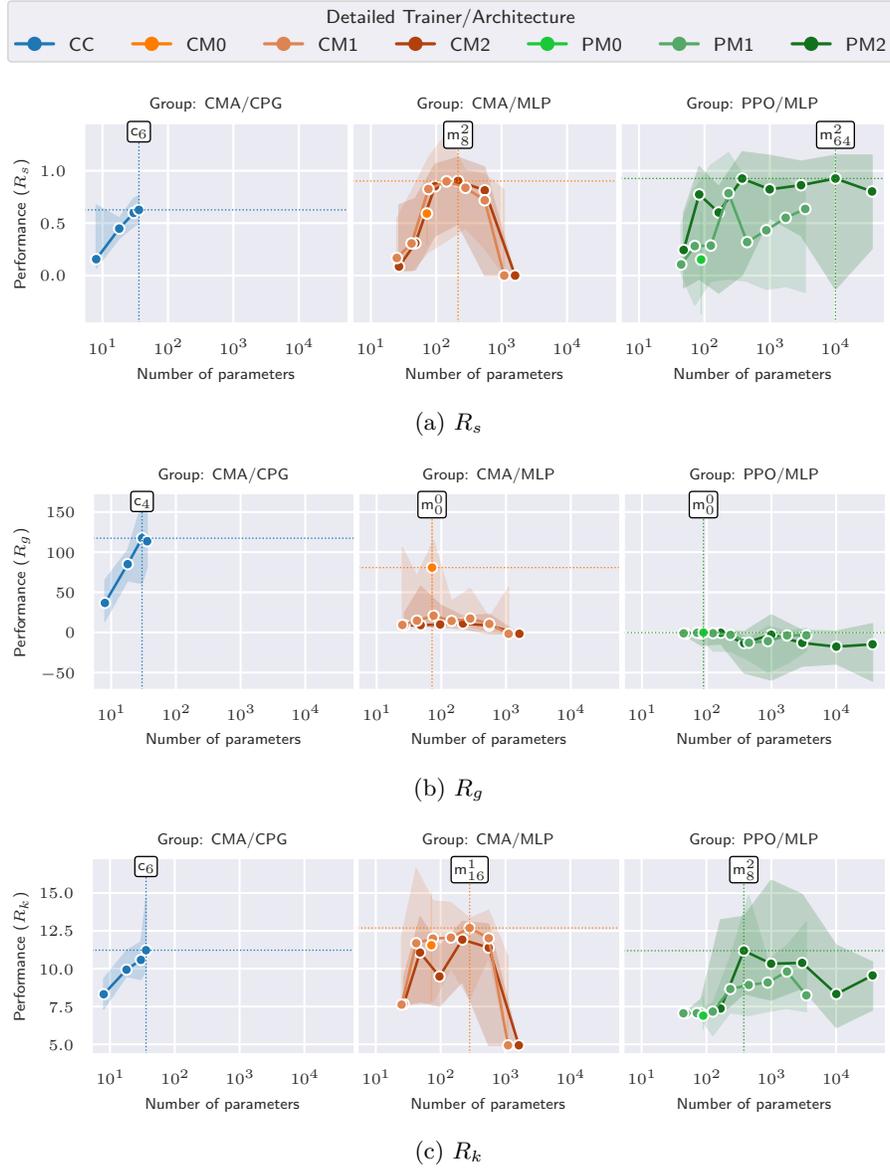}%
   \ifnum\i>1
    \vspace{-.25\baselineskip}
    \caption{\c}
    \label{fig:perf:relplot:\c}
    \vspace{.25\baselineskip}
   \fi%
  \end{subfigure}%
  
 }
 \caption{
  Relationship between performance and number of parameters for all configurations and three reward functions. CPGs trained with CMA-ES are displayed in blue, labeled CC. MLP trained with CMA-ES and PPO are labeled CM$i$ and PM$i$, respectively, for a depth of $d$. Sub-architectures with the highest median are annotated in each group and projections are provided for comparison. Trends show comparable performance on \rewK{}, MLP dominance for \rewS{} and CPG dominance for \rewG{}. Generally, bigger CPGs and smaller MLPs perform better.
 }
 \label{fig:perf:relplot}
\end{figure}

To discriminate between the different sub-architectures we first look at the raw performance on each reward function. 
As summarised in \autoref{fig:perf:relplot}, each training regime reacted very differently with each architecture, with the exception of CPGs (CC), which generally benefits from having more parameters.

MLPs trained with CMA-ES, instead, exhibit a bell-shaped curve for both \rewS{} and \rewK{}.
While using a version of CMA-ES more suited to large search spaces \cite{Varelas2018} could improve scalability, this setup shows what can be achieved with off-the-shelf solutions.
Furthermore, as the performance of PPO is either on par (\rewS{}) or lower (\rewG{}, \rewK{}) than that of CMA-ES, we can reasonably hypothesise that the parameter space is already explored close enough to near-optimal performance.

Indeed, one can see on \autoref{fig:perf:relplot:\rewS} that the best performing architecture for MLP is either $m^2_8$ (2 hidden layers of 8 neurons each) or $m^2_{64}$ depending on whether the training was performed with CMA-ES or PPO.
However, increasing the parameters by a factor of almost 50 only changed the performance by 3\%.
With respect to \rewK, the general trend is similar but with an important difference on PPO saturation: with this reward function, the larger networks do not even reach the performance of the smaller ones.
Given the relative straightforwardness of the task, this goes to show how overparametrisation can be detrimental to the learning process, when working under a similar budget.

The case of the gymnasium-based fitness is particularly interesting as it should have been more beneficial to training ANNs with RL, given that it is what it was designed for.
However, one can see on \autoref{fig:perf:relplot:\rewG} that MLPs perform very poorly with only CMA-trained perceptrons achieving meaningful rewards.
While this comes across as a counter-intuitive result, one must recall that \rewG{} actively penalises large motions, making the initial learning steps that much harder.
This problem does not come into play as strongly when considering CPGs which, for better or worse, have a very hard time not generating rhythmic patterns.

However, to identify what architecture/trainer combination performs best for this locomotion task we now only look at the values obtained by the best sub-architecture.
As mentioned before, CPGs strongly benefit from having more parameters and, with the exception of \rewG{}, should be used in a fully connected setting.
MLPs, however, differ as CMA-ES favors small networks ($m^2_8$ and $m^1_2$ for \rewS{} and \rewK{}, respectively) while PPO is more dispersed ($m^2_{64}$ and $m^2_8$ for \rewS{} and \rewK{}).
Interestingly, the gymnasium-based reward led to a dominance of pure Perceptrons ($m_0^0$) in both cases.

Performance-wise, MLPs are valid choices with both training methods when aiming for pure speed but do not strongly outperform CPGs\footnote{$R_s$: non significant via Mann-Whitney, Boneferroni correction}.
Conversely, the counter-intuitive shaping of \rewG{} made it so that frugal controllers are better modelled by CPGs or pure perceptrons\footnote{$R_g$: PPO/MLP lower than both CMA/CPG and CMA/MLP with p < 0.001}.
Finally, when targeting a more balanced gait, as promoted by \rewK{}, performance is not enough to make an informed choice (no statistical difference).
All results stated here are summarised in \autoref{tab:summary:perf} for easier comparison.

\subsection{Parameter Impact}

\def\impS{$I_s$}
\def\impG{$I_g$}
\def\impK{$I_k$}

In the context of robotics, raw task performance may not be the only metric by which policies are compared.
Indeed, efficiency may also play an important role, whether in terms of energy consumption, battery longevity or, as it interests us here, the parameter space.
As we have seen previously, a large parameter space does not correlate with better performance.
This links back to the low dimensionality of our observation and actions spaces, making it possible to have prohibitively high numbers of parameters.

To better capture the relationship between parameter count and performance and to empirically measure overparametrization instead of defining it a priori, we consider the Parameter Impact $I_F(f, p)$ as:
\begin{equation}%
 I_F(f, p) = \hat f / log_{10}(p)
\end{equation}%
where $p$ is the number of parameter in the policy and $\hat f$ is the normalised performance, obtained by uniformly scaling $f$ with respect to the performance distribution of individuals also trained with reward function $F$.
The use of logarithmic scaling ensures that the exponentially increasing parameter count of MLPs, especially in conjunction with PPO, do not overshadow actual performance and is consistent with pareto-front observations from the companion data \cite{kevin_2026_19633625}.
Given the three reward functions previously introduced, we have as many Parameter Impact metrics \impS{}, \impG{}, \impK{} normalising the performance of \rewS{}, \rewG{} and \rewK{}, respectively.
Thus performances are transformed into ratios, answering the question: relative to the global population, how much does the chosen parametrization help to get a good performance?

\begin{figure}
 \foreach \i/\c in {5, 7/\impS, 8/\impG, 6/\impK} {
  \begin{subfigure}{.99\textwidth}
   \centering
   \includegraphics[width=\textwidth, page=\i]{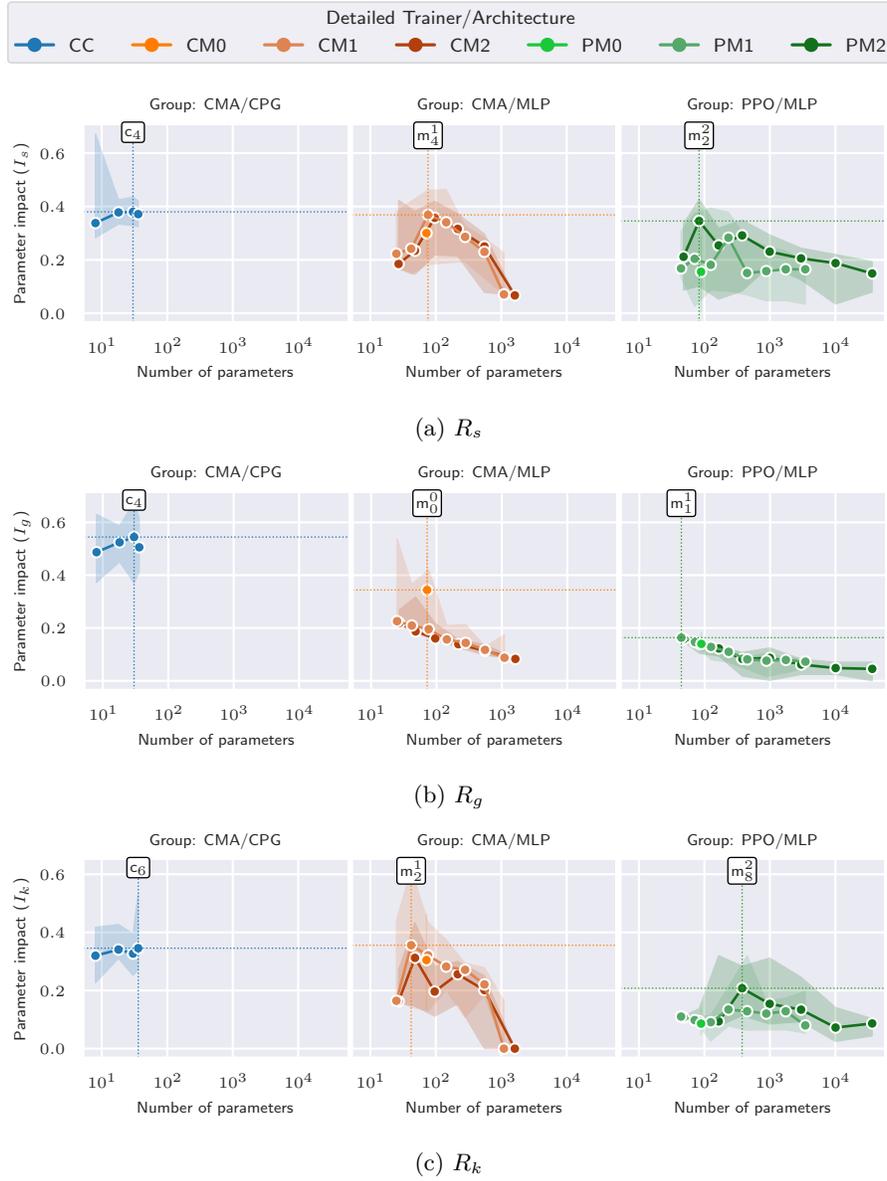}
   \ifnum
    \i>5\caption{\c}
    \label{fig:pi:relplot:\c}
   \fi
  \end{subfigure}
 }
 \caption{
  Parameter Impact across all configurations and reward functions.
  Color scheme is identical as \autoref{fig:perf:relplot}.
  Sub-architectures with the highest median are also annotated.
  Trends show comparable performance on \impS{}, CPG dominance for \impG{} and CMA-ES dominance for \impK{}.
  CPGs with almost fully connection topologies ($c_4$) and very small MLPs architectures stand out.
 }
 \label{fig:pi:relplot}
\end{figure}

Applying this metric to our dataset results in \autoref{fig:pi:relplot}, where one can better compare how much increasing parameters benefits performance, even across reward functions.
As the Parameter Impact only acts as a logarithmic scaling, the performances are qualitatively similar but their relationships across groups and reward schemes are not.

Indeed, the trends that were already visible with the raw performance are more easily accessible in this scaled down format.
Notably, it becomes clear that the 47\% increase in \rewS{} between CPGs and PPO-trained MLP is made with a parameter increase of 27958\% (10065 versus 36).
When scaled down with \impS{}, one can see that the latter was marked as less than half the efficiency of the former.
Overall, this leads to changes in the best performing architectures with $c_4$ taking the lead for the CPGs with its almost-fully connected topology.
On the one hand, MLPs trained with CMA-ES favour even smaller topologies, except for \impG which was only solved by perceptrons.
On the other hand, PPO produced more efficient networks at low scales with best performing architectures being $m^2_2, m^1_1, m^2_8$ for \impS{}, \impG and \impK, respectively.

As before, we summarise these results in \autoref{tab:summary:perf} and \cite{kevin_2026_19633625}.
Similarly to performance results, no architecture consistently outperforms all alternatives.
\impG{} favours CPGs\footnote{$I_g$: PPO/MLP < CMA/MLP < CMA/CPG, significant at $p < 0.01$} while \impK{} only shows an advantage for CMA-based training\footnote{$I_k$: PPO/MLP < CMA/CPG, $p < 0.001$ and PPO/MLP < CMA/MLP, $p < 0.05$} and \impS{} results in similar scores (no statistical difference).
Overall, evaluating on efficiency highlights the negative impact of increased parameter count in cases, such as this one, where performance is bounded.

\subsection{Cross-performance}

In addition to the relative performance of the various architectures on multiple reward functions, one could wonder whether any of the former have particular transferability to any other one of the latter.
We thus test the cross-performance of the various architectures to investigate if any are better at exhibiting a range of interesting behaviour, even when not explicitly selecting for them.

\begin{table}[t]
 \def\cpg{\color{blue}}
 \def\cmlp{\color{orange!87!black}}
 \def\pmlp{\color{green!75!black}}
 \fontsize{7.5}{10.5}\selectfont
 \caption{Left: overview of every architectures (CPG, MLP) and trainer (CMA, PPO) for each performance metric (Speed, \rewS; Frugality, \rewG; Stability, \rewK) and Parameter Impact (\impS, \impG, \impK). In each row, the performance is reported first and the champion sub-architecture is noted below. Best performing architecture/trainer pairs are highlighted in coloured bold. Stars denote rare occurrences where an architecture is significantly better than all alternatives. Right: cross-performance when trained on X but evaluated on Y, denoted $\bm{X}/Y$.}
 \vspace{.5\baselineskip}
 \begin{subtable}[b]{.49\textwidth}
  \centering%
  \phantomsubcaption\label{tab:summary:perf}%
  \begin{tabular}[b]{llllllll}
 &  & $R_s$ & $R_g$ & $R_k$ & $I_s$ & $I_g$ & $I_k$ \\
\midrule
\multirow[c]{4}{*}{CMA} & \multirow[c]{2}{*}{\cpg\bfseries{CPG}} & 0.63 & \cpg\bfseries{117.34} & 11.22 & \cpg\bfseries{0.38} & \cpg\bfseries{0.54} & 0.35 \\
 &  & c$_{6}$ & \cpg\bfseries{c$_{4}$}* & c$_{6}$ & \cpg\bfseries{c$_{4}$} & \cpg\bfseries{c$_{4}$}* & c$_{6}$ \\[.5em]
 & \multirow[c]{2}{*}{\cmlp\bfseries{MLP}} & 0.90 & 80.78 & \cmlp\bfseries{12.69} & 0.37 & 0.34 & \cmlp\bfseries{0.36} \\
 &  & m$^{2}_{8}$ & m$^{0}_{0}$ & \cmlp\bfseries{m$^{1}_{16}$} & m$^{1}_{4}$ & m$^{0}_{0}$ & \cmlp\bfseries{m$^{1}_{2}$} \\[.5em]
\multirow[c]{2}{*}{PPO} & \multirow[c]{2}{*}{\pmlp\bfseries{MLP}} & \pmlp\bfseries{0.93} & -0.23 & 11.19 & 0.35 & 0.16 & 0.21 \\
 &  & \pmlp\bfseries{m$^{2}_{64}$} & m$^{0}_{0}$ & m$^{2}_{8}$ & m$^{2}_{2}$ & m$^{1}_{1}$ & m$^{2}_{8}$ \\
\bottomrule
\end{tabular}
 \end{subtable}%
 \hfill%
 \begin{subtable}[b]{.49\textwidth}%
  \centering%
  \phantomsubcaption\label{tab:summary:cross_perf}%
  \begin{tabular}[b]{cccccc}
$\bm{R_s}/R_g$ & $\bm{R_s}/R_k$ & $\bm{R_g}/R_s$ & $\bm{R_g}/R_k$ & $\bm{R_k}/R_s$ & $\bm{R_k}/R_g$ \\
\midrule
\cpg\bfseries{0.64} & 0.25 & \cpg\bfseries{113.81} & \cpg\bfseries{42.78} & 7.81 & 6.99 \\
\cpg\bfseries{c$_{4}$} & c$_{6}$ & \cpg\bfseries{c$_{6}$} & \cpg\bfseries{c$_{6}$} & c$_{2}$ & c$_{0}$ \\[.5em]
0.55 & 0.26 & 86.02 & 17.13 & \cmlp\bfseries{7.89} & \cmlp\bfseries{7.01} \\
m$^{0}_{0}$ & m$^{1}_{32}$ & m$^{1}_{4}$ & m$^{0}_{0}$ & \cmlp\bfseries{m$^{2}_{2}$} & \cmlp\bfseries{m$^{1}_{8}$} \\[.5em]
0.02 & \pmlp\bfseries{0.31} & 97.69 & 31.27 & 6.98 & 5.89 \\
m$^{2}_{128}$ & \pmlp\bfseries{m$^{2}_{8}$} & m$^{2}_{8}$ & m$^{2}_{8}$ & m$^{2}_{1}$ & m$^{1}_{32}$ \\
\bottomrule
\end{tabular}
 \end{subtable}%
\end{table}

\autoref{tab:summary:cross_perf} provides an overview of the best performance of a given architecture when evaluated with a different reward function than the one it has been trained with.
From it, we can gather that, overall, CPGs seem to perform better across the board especially with respect to \rewG{}.
Indeed, MLPs converge towards very limited ranges of motion, leading to similarly bad performance at both moving forward (\rewS{}) or adopting a stable gait (\rewK{}).

Complementarily, we can also observe that training directly for stable gait has interesting side effect when paired with CMA-ES, as it also provides individuals with slightly better performance over the remaining two functions.
This however, is compounded by the relative proximity of CPGs on those same functions.

Thus, while CPGs do not clearly stand out in terms of pure performance, they do exhibit an overall better ranking and often statistically outperform MLPs with either CMA or PPO training.
This, as we stated before, is to be linked with built-in bias toward rhythmic patterns which, besides favouring locomotion, are indirectly rewarded in all considered fitness functions.

\begin{figure}[t]
 \centering
 \includegraphics[width=.75\textwidth]{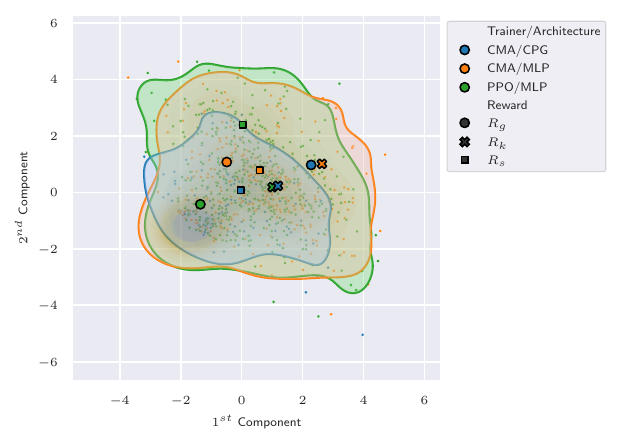}
 \caption{PCA of the 16D diversity space obtained by fitting sinusoidals onto each of the robot's foot. Explained variance ratios are 25.6\% and 22.8\% and contours cover 95\% of the distribution. Overall best performing individuals are highlighted to show the empirical diversity.}
 \label{fig:diversity}
\end{figure}

\subsection{Diversity}

Beyond performance, whether scaled or not, we finally address a potentially crucial component of the various architectures: their potential for producing diverse gaits.
To do so, we monitor the position of each of the robot's extremum brick.
Assuming a smooth gait, these should also follow a rhythmic pattern with each ``foot'' being raised in turn to maximise locomotion, as is seen in biological gaits \cite{bhattasali_neural_2024}.
However, as solutions have not been constrained to follow such a rule we instead model feet independently as a sinusoidal of the form:

\begin{equation}
 f_i(t) = A_i sin(\omega_i t + \phi_i) + c_i.
\end{equation}

This is then fitted to the actual trajectory of foot $i$ using the scipy library to provide estimated parameters.
To remove dubious fits, we constrain $A_i < 0.5$, $c_i < 0.5$ and $|\frac{w_i}{2\pi}| < 0.25$, and set all parameters to 0 whenever any constraint is not met.
This prevents immobile feet from being associated with curves with very high parameters.

Given that the robot has four feet, this makes it possible to represent every individual in a 16-dimensional space describing their intrinsic gait (or lack thereof).
This space has then been reduced to 2 dimensions via Principal Component Analysis and plotted in \autoref{fig:diversity}.
While the explained variance ratios are admittedly low (25.6\% and 22.8\%), this provides a measure of understanding into how diverse the solutions are, given a specific architecture/trainer pair.
We can see that both MLP variations occupy a larger area.
While this does not come at a surprise, given that CPGs have fewer parameters and a built-in tendency towards specific patterns, we here provide an empirical demonstration of just such a behaviour.
Whether or not this is a limitation, however, is debatable.
Indeed, even though CPGs are less capable of generating as wide a range of gait as MLPs, this same limitation also leads to higher overall cross-performance.

However, the curious reader can also look at videos\footnote{\resizebox{.95\textwidth}{!}{\url{https://www.youtube.com/playlist?list=PLP2arCAfRu39f4zFYQamZWD9hrMH57HBh}}} of the champions highlighted on \autoref{fig:diversity}, to further inform their choice between performance, efficiency and diversity.

\section{Discussion}

In this work, we systematically compare the performance of 17 variations of different neural architectures (CPGs and MLPs), with 2 training algorithms (CMA-ES and PPO), on 3 reward functions favouring different gaits (fast, frugal, stable) for a fixed quadrupedal, so-called "spider", morphology.
These comparisons were performed using off-the-shelf state-of-the-art algorithms to provide an easily replicable context which future experimenters could compare with.

Results show that, overall, low parameter counts led to better performance especially when aiming for more balanced behaviour (frugal, stable) than exploitive ones (speed).
CPGs showed a marked preference for the upper ranges of their parameter space with classes $c_4$ (30 parameters) and $c_6$ (36) outperforming loosely connected alternatives.
MLPs performed best with very shallow networks as champion architectures use between 0 and 16 hidden neurons, in either one or two layers.

In terms of training algorithm, CMA-ES had better all-around outcomes, in part thanks to its gradient-free approach which made it possible to train CPGs and to save on the parameter cost of a critic.
Indeed, when looking at networks' efficiency via our Parameter Impact metric, we see that shallow MLPs and CPGs strike the optimal balance between expressivity and performance.
However, that very same low parameter count also make CPGs less proficient as generating diverse gaits, as observed when looking at the distribution of fitted sinusoidal parameters.
While there are no current off-the-shelf solutions for CPG training with RL methods, the marked interest from the research community shows that this might change in the near future.
An extension of this study to encompass this fourth alternative would further demonstrate whether CPGs are an efficient, low-parameter, paradigm.
Additionally, the soundness of these results will be confirmed on different morphologies both within the same platform and without.

\begin{credits}
\subsubsection{\ackname} 
This research was funded by the Hybrid Intelligence Center, a 10-year programme funded by the Dutch Ministry of Education, Culture and Science through the Netherlands Organisation for Scientific Research, \newline\url{https://hybrid-intelligence-centre.nl}, grant number 024.004.022.

\subsubsection{\discintname}
The authors have no competing interests to declare that are relevant to the content of this article. 
\end{credits}

\bibliographystyle{splncs04}
\bibliography{autorefs}

\end{document}